\documentclass[lettersize,journal]{IEEEtran}
\usepackage{amsmath,amsfonts}
\usepackage{algorithmic}
\usepackage{algorithm}
\usepackage{array}
\usepackage[caption=false,font=normalsize,labelfont=sf,textfont=sf]{subfig}
\usepackage{textcomp}
\usepackage{stfloats}
\usepackage{url}
\usepackage{verbatim}
\usepackage{graphicx}
\usepackage{tablefootnote}
\def\BibTeX{{\rm B\kern-.05em{\sc i\kern-.025em b}\kern-.08em
    T\kern-.1667em\lower.7ex\hbox{E}\kern-.125emX}}
\usepackage{balance}
\begin{document}
\title{MKANet: A Lightweight Network with Sobel Boundary Loss for Efficient Land-cover Classification of Satellite Remote Sensing Imagery}

\author{Zhiqi Zhang, Wen Lu, Jinshan Cao, Guangqi Xie
\thanks{Zhiqi Zhang, Wen Lu and Jinshan Cao are with School of Computer Science, Hubei University of Technology, Wuhan 430068, China (e-mail: zzq540@hbut.edu.cn; wenlu@hbut.edu.cn; caojs@hbut.edu.cn).}
\thanks{Zhiqi Zhang and Guangqi Xie are with State Key Laboratory of Information Engineering in Surveying, Mapping, and Remote Sensing, Wuhan University, Wuhan 430079, China (e-mail: xiegqrs@whu.edu.cn).}
\thanks{Zhiqi Zhang and Wen Lu contributed equally to this work, and Jinshan Cao is the corresponding author.}
}

\markboth{}%
{MKANet: A Lightweight Network with Boundary Loss for Efficient Land-cover Classification of Satellite Remote Sensing Imagery}

\maketitle

\begin{abstract}
Land cover classification is a multi-class segmentation task to classify each pixel into a certain natural or man-made category of the earth surface, such as water, soil, natural vegetation, crops, and human infrastructure. Limited by hardware computational resources and memory capacity, most existing studies preprocessed original remote sensing images by down sampling or cropping them into small patches less than 512×512 pixels before sending them to a deep neural network. However, down sampling images incurs spatial detail loss, renders small segments hard to discriminate, and reverses the spatial resolution progress obtained by decades of years of efforts. Cropping images into small patches causes a loss of long-range context information, and restoring the predicted results to their original size brings extra latency. In response to the above weaknesses, we present an efficient lightweight semantic segmentation network termed MKANet. Aimed at the characteristics of top view high-resolution remote sensing imagery, MKANet utilizes sharing kernels to simultaneously and equally handle ground segments of inconsistent scales, and also employs parallel and shallow architecture to boost inference speed and friendly support image patches more than 10X larger. To enhance boundary and small segments discrimination, we also propose a method that captures category impurity areas, exploits boundary information and exerts an extra penalty on boundaries and small segment misjudgment. Both visual interpretations and quantitative metrics of extensive experiments demonstrate that MKANet acquires state-of-the-art accuracy on two land-cover classification datasets and infers 2X faster than other competitive lightweight networks. All these merits highlight the potential of MKANet in practical applications. 
\end{abstract}


\section{Introduction}
\IEEEPARstart{N}{owadays}, various satellite constellations with shorter revisit period and wider observation coverage have formed the global earth observation system which can quickly obtain huge amounts of high spatial resolution, high temporal resolution and high spectral resolution remote sensing imagery. For example, China has 30 to 50 high-resolution remote sensing satellites in orbit, by conservative estimation, several hundreds TB data is acquired every day \cite{li2017earth}. However, with regard to the acquisition speed, rapid intelligent processing of remote sensing data still lags. In the new era of artificial intelligence, how to realize instant perception and cognition of remote sensing imagery has become an urgent problem to be solved.

\begin{figure}[!t]
\centering
\includegraphics[width=3.4in]{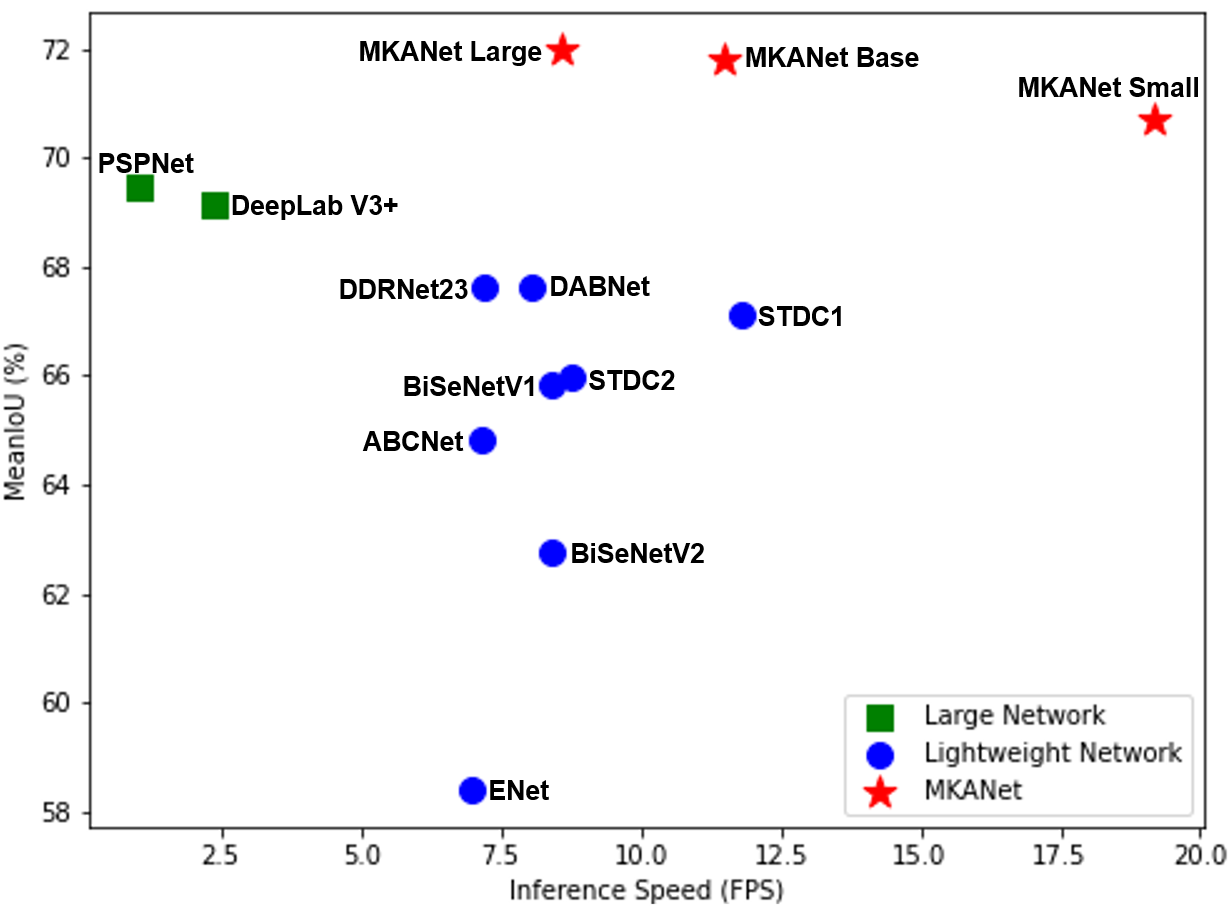}
\caption{Speed-Accuracy performance comparison on the DeepGlobe Land Cover dataset, the proposed MKANets achieve state-of-the-art speed-accuracy trade-off.}
\label{DeepGlobeSpeedAccuracy}
\end{figure}


Land cover classification is a multi-class segmentation task to classify each pixel into a certain natural or man-made category of the earth surface, such as water, soil, natural vegetation, crops, and human infrastructure. The land cover and its change influence the ecosystem, human health, society development and economic growth. Last several decades of years have witnessed the improvement of spatial resolution of remote sensing imagery from 30m to sub-meter. With richer details and structural information of objects emerging in remote sensing imagery, land cover classification methods have shifted from discriminating in spectral or spectral-spatial information of local pixels to extracting contextual information and spatial relationship of ground objects \cite{tong2020land}. Among them, deep neural networks (DNN) have been widely used for their strong feature extraction and high-level semantic modelling ability. However, large computational resources consumpution brings slow inference speed and restricts the practical application of DNN in remote sensing imagery. Meanwhile, incapability of processing large size image patches causes cropping size too small, the resulted loss of long-range context information is detrimental to prediction accuracy.

To acquire high accuracy, conventional semantic segmentation networks, such as UNet \cite{ronneberger2015u}, FC-DenseNet \cite{jegou2017one}, and DeepLabv3+ \cite{chen2018encoder}, usually adopt a wide and deep backbone as an encoder at the cost of large computational complexity and memory occupation. In the task of land-cover classification, limited by GPU memory capacity, most existing studies preprocessed the original remote sensing images by down sampling or cropping them into small patches less than 512×512 pixels before sending them to a deep neural network. For example, CFAMNet \cite{wang2022semantic} proposes a class feature attention mechanism fused with an improved Deeplabv3+ network. To avoid memory overflow, 150 remote sensing images of 7200×6800 pixels were cropped into 20776 images of 128×128 pixels. DEANet \cite{wei2021deanet} proposes a dual-branch encoder structure that depends on VGGNet \cite{simonyan2014very} or ResNet \cite{he2016deep}, in their experiments, each image with a resolution of 2448×2448 pixels was compressed to half the size and then divided into subimages with a resolution of 512×512 pixels. DISNet \cite{zhang2021remote} integrates the dual attention mechanism module, including the spatial attention mechanism and channel attention mechanism, into the Deeplabv3+ network. In their experiments, the original images were also cropped into small patches of 512×512 pixels before being sent into the network. 

However, it takes decades of years of efforts to improve spatial resolution of remote sensing imagery, however, down sampling reverses this progress and incurs spatial detail loss. The rich details of objects, such as the geometrical shape and structural content of objects, are blurred by down sampling. It renders small segments hard to discriminate, thus offsetting the gain enabled by the large backbone. 

\begin{figure}[!t]
\centering
\includegraphics[width=3.4in]{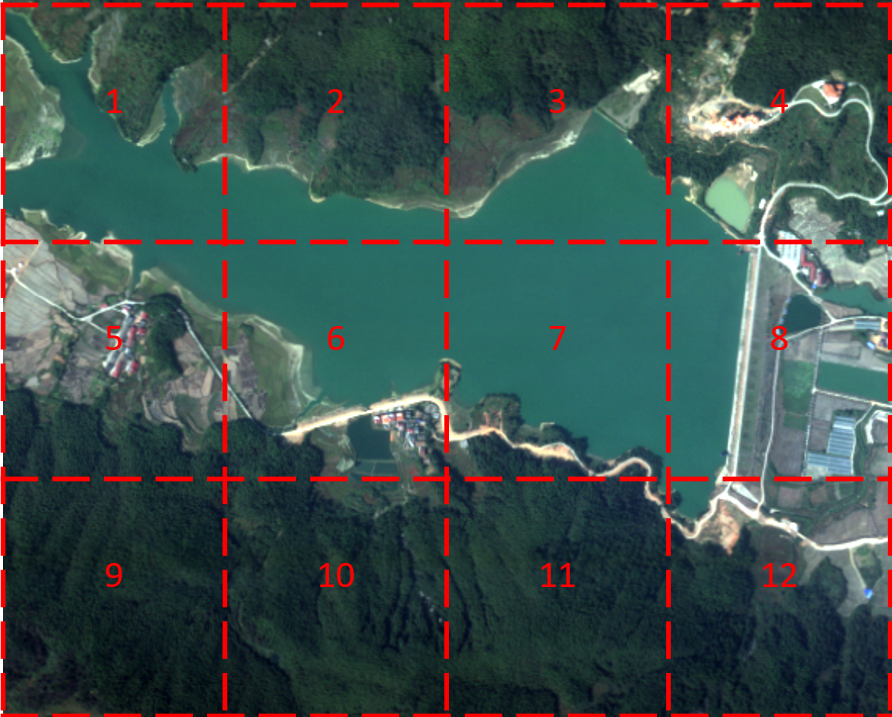}
\caption{A romote sensing image with a resolution of 2048×1536 pixels is cropped into 12 small patches with a resolution of 512×512 pixels.}
\label{Patches}
\end{figure}

On the other hand, cropping the original images into small patches as less than or equal to 512×512 pixels causes a loss of long-range context information and leads to misjudgment. As shown in Figure \ref{Patches}, a romote sensing image with a resolution of 2048×1536 pixels is cropped into 12 small patches with a resolution of 512×512 pixels. If view the whole original image, it is clear that the water surface is a lake; however, if view the individual small patches, the water surface may be misjudged as a river. Therefore, compared with images in other domains, such as street view images in autonomous driving field, support of large size image input is more important for the correct semantic segmentation of remote sensing images (In Section \ref{ablation various cropping sizes}, we investigate the influence of input image size on prediction accuracy, and demonstrate that the loss of long-range context information caused by small cropping size would incur misjudgement and lower accuracy). Another drawback of cropping is that cropping the images and restoring the predicted results to their original size would incur extra latency. Hence, aimed at the characteristics of top view high-resolution remote sensing imagery, it is necessary to redesign the architecture of semantic segmentation networks to support large size image patches.

As presented in Figure \ref{Patches}, the abundant small segments, rich boundaries and small interclass variance in remote sensing images are all likely to cause semantic ambiguity near the boundaries and small segments. Meanwhile, areas where exist multiple land-cover categories contain richer information and are more prone to be misjudged. In the other aspect, the number of interior pixels grows quadratically with segment size and can far exceed the number of boundary pixels, which only grows linearly. However, the ground truth masks and conventional loss functions value all pixels equally and is less sensitive to boundary quality. Hence, it is necessary to capture category impurity areas and implement an effective measure to reinject boundary information into the semantic segmentation network.

In summary, slow inference speed, incapability of processing large size image patches, and easy misjudgement on boundaries and small segments are three factors that restrict the practical applications of semantic segmentation networks. To alleviate these three problems, we present an efficient lightweight semantic segmentation network termed Multi-branch Kernel-sharing Atrous convolution network (MKANet) and propose the Sobel Boundary Loss for efficient and accurate land-cover classification of remote sensing imagery. The contributions of this paper can be summarized in three aspects:

\begin{enumerate}

\item{Aimed at the characteristics of top view remote sensing imagery, we handcraft the Multi-branch Kernel-sharing Atrous convolution (MKA) module for multi-scale feature extraction.}

\item{For large input image size support and fast inference speed, we design a shallow semantic segmentation network (MKANet) based on MKA modules.}

\item{For accurate prediction on boundaries and small segments, we propose a novel boundary loss named Sobel Boundary Loss.}

\end{enumerate}

\section{Related Work}
\noindent In this section, we first present some representative lightweight semantic segmentation networks and our improvement direction; then we introduce the original kernel-sharing mechanism as well as our reasons to adopt it, and its limitation.

\subsection{Lightweight Semantic Segmentation Networks} \label{section:lightweight_networks} 
\noindent Certain semantic segmentation networks employ lightweight backbones, so compared with large networks, they consume fewer hardware resources. For example, ENet \cite{paszke2016enet} employs an early downsampling strategy and asymmetric architecture that consists of a large encoder and a small decoder in pursuit of real-time processing. BiSeNetV1 \cite{yu2018bisenet} and BiSeNetV2 \cite{yu2021bisenet} propose a two-pathway architecture: the first pathway captures the spatial details with wide channels and shallow layers, and the second pathway extracts the categorical semantics with narrow channels and deep layers and then fuses the output features of these two paths to make the final prediction. Hong et al. proposed a deep dual-resolution network (DDRNet) \cite{hong2021deep} that consists of two deep branches between which multiple bilateral fusions are performed. To guarantee accuracy without drastically increasing computational consumption, ABCNet \cite{li2021abcnet} replaces the dot-product attention mechanism of quadratic complexity with a linear attention mechanism for global contextual information extraction. DABNet \cite{li2019dabnet} adopts depthwise asymmetric convolution and dilated convolution to build a bottleneck structure for parameter reduction. DFNet \cite{li2019partial} utilizes the partial order pruning algorithm to obtain a lightweight backbone and efficient decoder. 

Although these lightweight networks are computationally inexpensive, there is still some gap in accuracy between them and large networks. Empirical observation shows that prediction errors are more likely to occur on boundaries and small segments \cite{yuan2020segfix}, this observation prompts us to propose a novel boundary loss as detailed in Section \ref{Sobel Boundary Loss} to bridge the accuracy gap between lightweight networks and large networks.

\subsection{Kernel-Sharing Mechanism} \label{section:Kernel-Sharing_Mechanism} 
\noindent The authors of KSAC \cite{huang2021see} argued the weakness of the original ASPP \cite{chen2017rethinking} structure is that the kernels in the branch with small atrous rates only learn details and handle small objects well, while the kernels in the branch with large atrous rates are only able to extract features with large receptive fields. The lack of communication among branches compromises the generalizability of individual kernels. To tackle this problem, they proposed that multiple branches with different atrous rates can share a single kernel, so the shared kernel is able to scan the input feature maps more than once with both small and large receptive fields. Another benefit is that the objects of various sizes can all contribute to the training of the shared kernel, so the number of effective training samples increases, the representation ability of the shared kernel is thus improved. KSAC adopts the architecture of DeepLabV3+, its modified ASPP structure consists of a 1×1 convolutional branch, a global average pooling branch followed by a 1×1 convolution, and three kernel-sharing atrous convolutional branches with rates (6, 12, 18).

The spatial resolutions of different satellites are various, which leads land objects of the same category have different scales. On the other hand, lands belonging to the same category have different areas. The feature of multi-scales in remote sensing images accords with the advantages of kernel-sharing mechanism, so we decided to adopt this mechanism in the basic module design.

However, direct introduction of KSAC to basic module is impossible, because KSAC can only be applied once as the last stage of the encoder. Firstly, its global average pooling branch is supposed to obtain the image-level features; secondly its large atrous rates (6, 12, 18) are not suitable for extracting low level features, especially for remote sensing images in which objects have smaller spatial scales than those in generous images. Therefore, constructing a backbone by purely stacking the original KSAC structure is not feasible. So, we handcrafte a novel multi-branch module as detailed in Section \ref{section:MKA}, the newly proposed module can be stacked in multiple stages as the backbone for semantic segmentation.

\section{Proposed Method}
\noindent In this section, we first introduce the MKA module which constitute the backbone of the network; then we show the network architecture that infers 2X faster than other competitive
lightweight networks; at last we present the Sobel Boundary Loss that helps boundary recovery and improves small segments discrimination.

\subsection{Multi-branch Kernel-sharing Atrous convolution Module} \label{section:MKA} 
\begin{figure}[!t]
\centering
\includegraphics[width=3.4in]{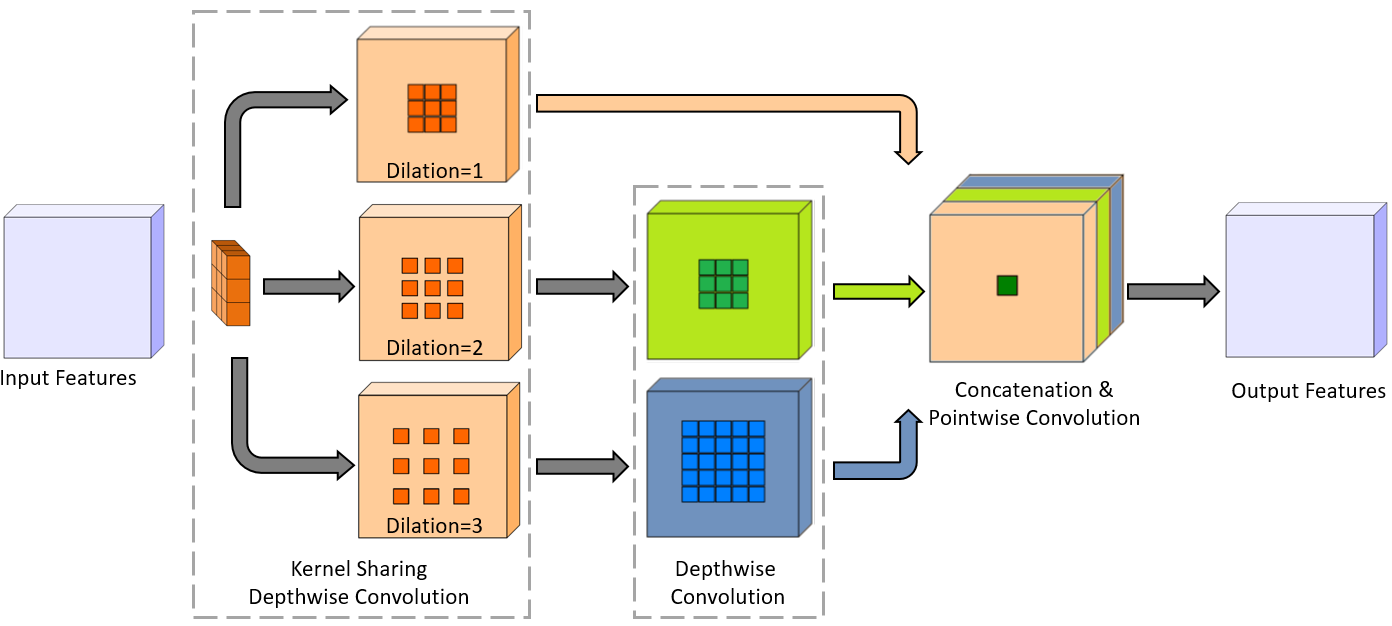}
\caption{Structure of the MKA module. The orange square represents the kernel shared by three depthwise atrous convolutional branches with dilation rates (1, 2, 3).}
\label{Structure_of_MKA_module}
\end{figure}

\noindent Conventional networks usually accumulate contextual information over large receptive fields by stacking a series of convolutional layers, so they have deep network architectures that consist of dozens of layers. Some networks even have more than one hundred layers, for example FC-DenseNet \cite{jegou2017one}. However, one of the costs of building a deep architecture is slow inference speed. For high efficency and fast inference speed, we design the Multi-branch Kernel-sharing Atrous convolution (MKA) module, as illustrated in Figure \ref{Structure_of_MKA_module}. Its parallel structure and kernel sharing mechanism can simultaneously capture a wider range of contexts for large segments and local detailed information for small segments and boundaries. Specifically, the receptive field of a typical three-branch MKA module equals that of five $3 \times 3$ convolutional layers connected in series. Different from ASPP or KSAC which can only be applied once as the last part of a backbone, MKA modules can be stacked in series as the backbone for semantic segmentation networks. Hence, with MKA modules, it is no longer necessary to build a deep network architecture. Meanwhile, the computation cost of a MKA module is inexpensive, along with its parallel structure, the MKA module can greatly boost inference speed. 

The MKA module consists of three parts:

\begin{itemize}
\item{Multibranch kernel-sharing depthwise atrous convolutions}
\item{Multibranch depthwise convolutions}
\item{Concatenation \& pointwise convolution}
\end{itemize}

\subsubsection{Part 1: Multibranch Kernel-sharing Depthwise Atrous Convolutions}
\noindent Assume that the number of channels of the input and output features is $N$ and that the number of branches is $M$. One $3 \times 3$ kernel is shared by the $M$ depthwise atrous convolutions with dilation rates of $1$, $2$, ... and $M$. Next, batch normalization is applied in each branch. 

Compared with KSAC \cite{huang2021see}, the MKA module abandons the $1 \times 1$ convolutional branch and the global average pooling branch. The dilation rates also decrease from (6, 12, 18) to (1, 2, 3). To further reduce computation complexity and memory occupation, regular atrous convolutions are replaced by depthwise atrous convolutions, decreasing the kernel parameters, computation cost, and memory footprint to $1/N$.

This design inherits the merits of the kernel-sharing mechanism. The generalization ability of the shared kernels is enhanced by learning both local detailed features of small segments and global semantical features of large segments. The kernel-sharing mechanism can also be considered feature augmentation performed inside the network, which is complementary to data augmentation performed in the preprocessing stage, to enhance the representation ability of kernels.

\subsubsection{Part 2: Multibranch Depthwise Convolutions}
\noindent Since atrous convolution introduces zeros in the convolutional kernel, within a kernel of size $k_d \times k_d$, the actual pixels that participate in the computation are just $k \times k$, with a gap of $r - 1$ between them. Hence, a kernel only views the feature map in a checkerboard fashion and loses a large portion of information. Furthermore, the adjacent points of its output feature map do not have any common pixels participating in the computation, thereby causing the output feature map to be unsmooth. This gridding artifact issue is exacerbated when atrous convolutions are stacked layer by layer. To alleviate this detrimental effect, for the $i$th branch ($i > 1$), a depthwise regular convolution with kernel size ($2 \times i - 1$) is added, followed by batch normalization. Note that atrous convolution with a dilation rate of 1 is just regular convolution; thus, for the first branch, nothing is added. Again, depthwise convolutions are applied here to reduce computation and memory costs. With the exception of smoothing the output feature maps of the preceding part, these depthwise convolutions can further extract useful information.

\subsubsection{Part 3: Concatenation \& Pointwise Convolution}
\noindent After the second part, the output features of each branch are concatenated, and then a $1 \times 1$ convolution is applied to the fused features. This part has two functions: generating new features through linear combinations and compressing the number of channels of the fused features from $M \times N$ to $N$ to reduce the computational complexity of the next module.

\subsubsection{Complexity Analysis}
\noindent The number of parameters in the first part is $3 \times 3 \times N = 9N$; in the second part is $((4M^3 - M) / 3 - 1) \times N$; and in the third part is $M \times N \times N = MN^2$. The number of branches $M$ is suggested to be 3, which is substantially less than the number of channels $N$. Hence, the total parameters of the MKA module are approximately $MN^2$, which is even less than the parameters of one regular $3 \times 3$ convolution.

\subsection{Network Architecture} \label{section:Network_Architecture}
\begin{figure*}[!t]
\centering
\includegraphics[width=6in]{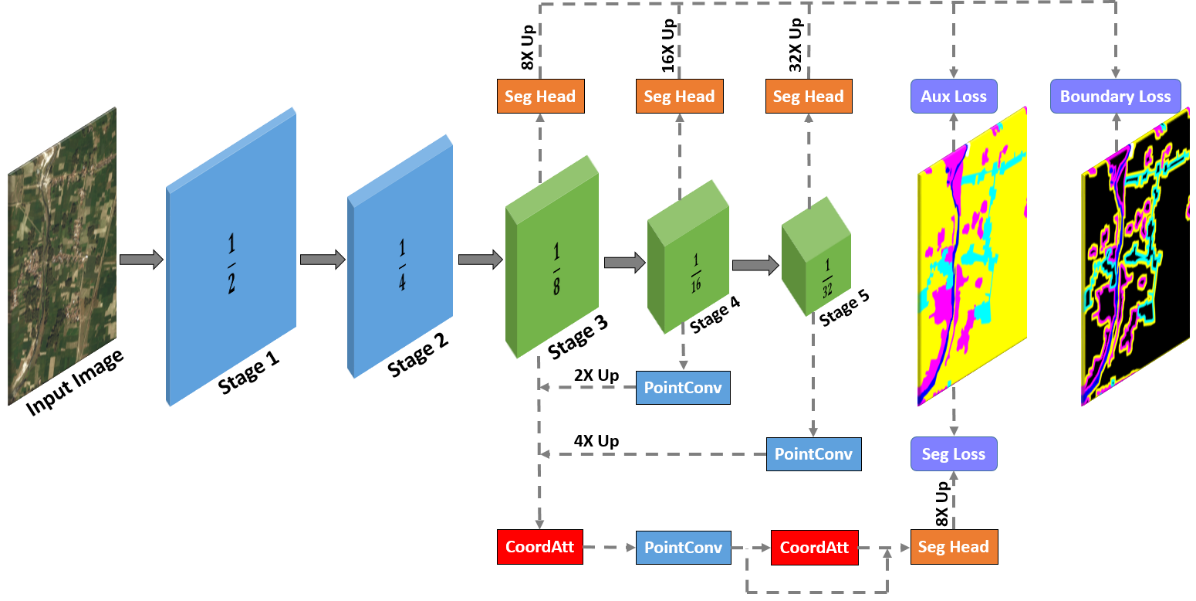}
\caption{Architecture of MKANet. The blue cuboid represents the $3 \times 3$ convolution with stride 2,
and the green cuboid represents $3 \times 3$ convolution with stride 2 plus one or more MKA modules.}
\label{Architecture_of_MKANet}
\end{figure*}

\noindent For faster inference speed and small memory occupation, based on MKA modules, we design a lightweight semantic segmentation network named MKANet, as illustrated in Figure \ref{Architecture_of_MKANet}. Attributed to the large receptive field of MKA modules, the network architecture of MKANet is very shallow. It consists of two initial convolutional layers and three MKA modules as the encoder and two Coordinate Attention Modules (CAM) \cite{hou2021coordinate} as the decoder to fuse multiscale feature maps from different stages. By horizontal and vertical extent pooling kernels, CAMs can capture long-range dependencies along one spatial direction and preserve precise positional information along the other spatial direction, thus more accurately augmenting the representations of the objects of interest in the fused feature maps.

As presented in Section \ref{Inference_Speed}, this shallow but effective architecture makes MKANet capable of supporting input image size more than 10 times larger than that supported by conventional networks. Furthermore, compared with other competitive lightweight networks, the inference speed of MKANet is twice faster.

\subsubsection{Encoder}
\noindent The encoder of MKANet has five stages, with each stage downsizing the feature maps by 2X. Its structure is detailed in Table \ref{tab_MKA}.

\begin{table}
\begin{center}
\caption{Encoder design.}
\label{tab_MKA}
\begin{tabular}{ c  c  c  c }
\hline
\bf{Stages} & \bf{Output Size} & \bf{Operation} & \bf{Output Channels} \\
\hline
Input Image & $2400 \times 2400$ &  & $3$\\

Stage 1 & $1200 \times 1200$ & ConvS2 & $c / 2$ \\

Stage 2 & $600 \times 600$ & ConvS2 & $c$ \\
\hline 
Stage 3 & $300 \times 300$ & ConvS2 & $c \times 2$ \\
   		 & $300 \times 300$ & MKA $\times$ $r$ & $c \times 2$ \\
\hline 
Stage 4 & $150 \times 150$ & ConvS2 & $c \times 4$ \\
   		 & $150 \times 150$ & MKA $\times$ $r$ & $c \times 4$ \\
\hline 
Stage 5 & $75 \times 75$ & ConvS2 & $c \times 8$ \\
   		 & $75 \times 75$ & MKA $\times$ $r$ & $c \times 8$ \\
\hline 
\end{tabular}
\end{center}
ConvS2: $3 \times 3$ convolution with stride $2$, batch normalization, ReLU activation. \\
$c$: the parameter controls the width of the backbone. \\
$r$: repeating time of MKA module.
\end{table}

Each stage begins with a $3 \times 3$ convolution of stride 2, followed by batch normalization and ReLU activation. MKA modules are then repeated $r$ times in each stage since Stage 3. $r$ controls the depth, while $c$ controls the width of the backbone. MKANet has 3 typical sizes: Small ($c=64, r = 1$), Base ($c=96, r = 1$) and Large ($c=128, r = 1$).

\subsubsection{Decoder}
\noindent The purpose of the first two stages is to extract simple, low-level features and to quickly downsize the resolution for computation reduction. Hence, the decoder only collects the deep context feature representations extracted by the MKA modules in Stages 3 to 5. 

\begin{figure}[!t]
\centering
\includegraphics[width=1.5in]{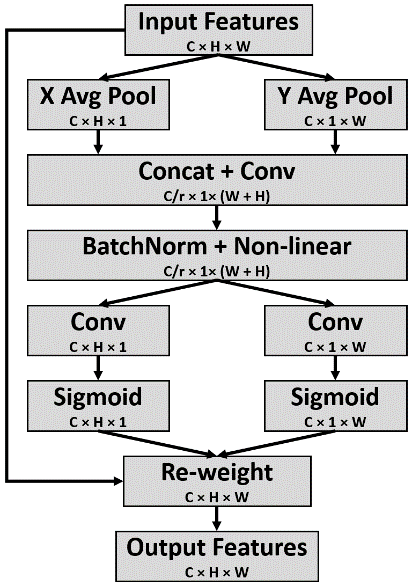}
\caption{Structure of Coordinate Attention Module (CAM). 'X Avg Pool' and 'Y Avg Pool' refer to 1D horizontal global pooling and 1D vertical global pooling, respectively.}
\label{CAM}
\end{figure}

To fuse multiscale feature maps, certain efficient networks, such as BiSeNet V1 \cite{yu2018bisenet} and STDC \cite{fan2021rethinking}, use squeeze-and-excitation (SE) attention \cite{hu2018squeeze} to transform a feature tensor to a single feature vector via 2D global pooling and rescale the feature maps to selectively strengthen the important feature maps and to weaken the useless feature maps. Although SE attention can raise the representation power of a network at a low computational cost, it only encodes interchannel information without embedding position-sensitive information, which may help locate the objects of interest. To embed positional information into channel attention, Hou et al. proposed the Coordinate Attention Modules (CAM), which utilizes two 1D global pooling operations to aggregate features along the horizontal and vertical directions so that the two generated, direction-aware feature maps can capture long-range dependency along one spatial direction and preserve precise positional information along the other spatial direction. The detailed structure of the CAM is shown in Figure \ref{CAM}.

To tell the network 'what' and 'where' to attend, two CAMs were employed to fuse the multiscale feature maps output by stages 3 to 5. Specifically, the feature maps output by Stage 4 and Stage 5 are compressed from $4c$ and $8c$, respectively, to $2c$ in the channel by pointwise convolution, up sampled by 2X and 4X, respectively, and concatenated with the feature maps output by Stage 3. After the feature maps are put through the first CAM to derive a combination of features with enhanced representation, a pointwise convolution is employed to promote communication of information among the channels and to further compress the number of channels from $6c$ to $2c$. The compressed feature maps pass the second CAM with the residual connection.

\subsection{Semantic Segmentation Losses}
\begin{figure}[!t]
\centering
\includegraphics[width=1in]{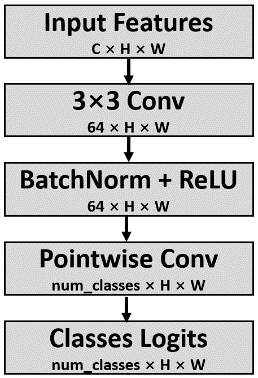}
\caption{Structure of the semantic segmentation head.}
\label{semantic_segmentation_head}
\end{figure}

\noindent The semantic segmentation head, as illustrated in Figure \ref{semantic_segmentation_head}, converts the output feature maps of the decoder into class logits, which are then up sampled by 8X to restore them to the same resolution as the input image. The up sampled class logits are compared with the ground truth by the main semantic segmentation loss function.

To enhance the feature extraction ability of the MKA modules, three auxiliary semantic segmentation heads are added on top of the output features of Stage 3 to Stage 5 in the training phase. In the inference phase, the three auxiliary heads are discarded, without additional computational cost. The output class logits of the three auxiliary semantic segmentation heads are up sampled 8X, 16X, and 32X before being sent to three auxiliary semantic segmentation loss functions and three Boundary Loss functions.

\subsection{Sobel Boundary Loss}\label{Sobel Boundary Loss} 
\noindent Sobel operator is a discrete differentiation gradient-based operator that computes the gradient approximation of image intensity for edge detection. It employs two 3 × 3 kernels to convolve with the input image to calculate the vertical and horizontal derivative approximations respectively.

\begin{figure*}[!t]
\centering
\includegraphics[width=5in]{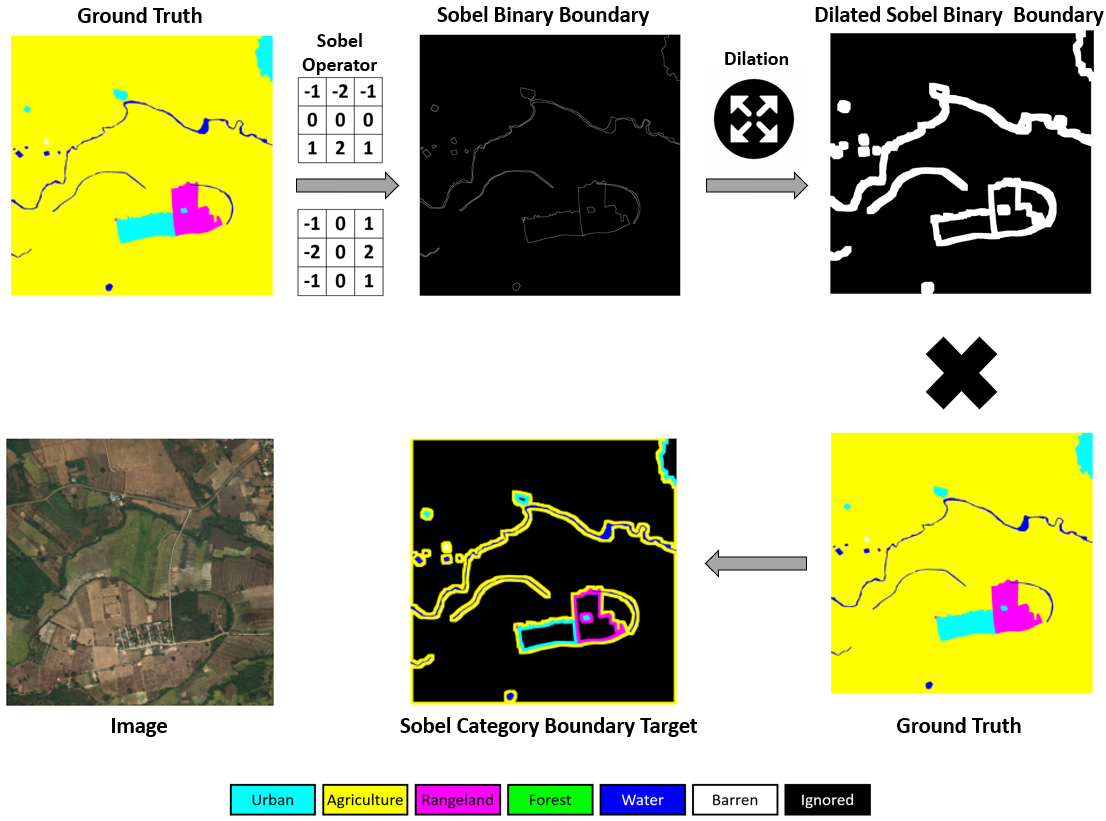}
\caption{The procedure of generating Sobel Category Boundary Target, with dilation rate $d = 50$ pixels.}
\label{Sobel_Boundary_Formation}
\end{figure*}

\begin{figure}[!t]
\centering
\includegraphics[width=3in]{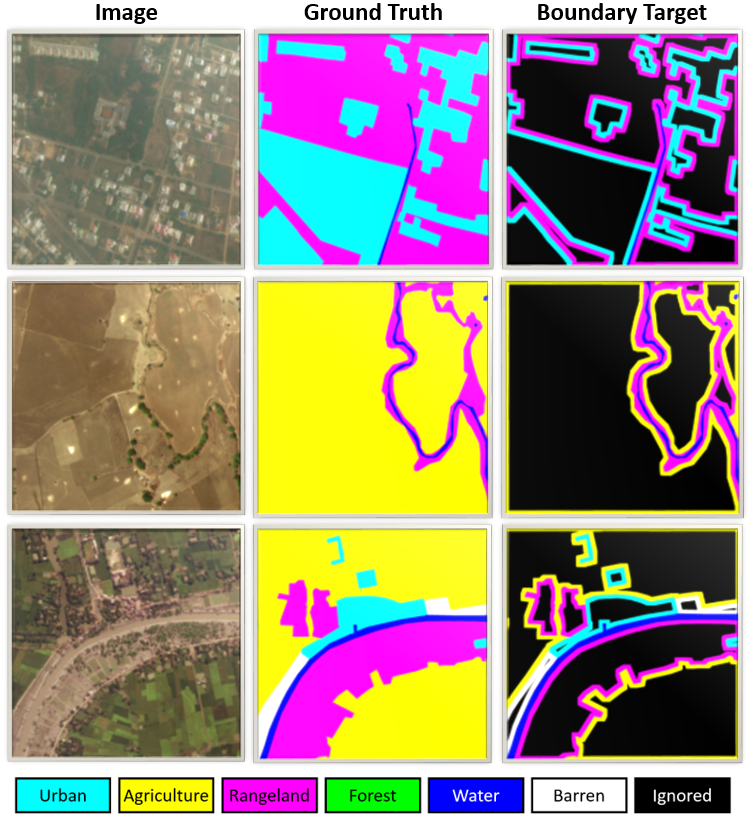}
\caption{Images, ground truths and boundary targets generated by Sobel kernels and dilation operation with $d = 50$ pixels.  }
\label{boundary_targets}
\end{figure}

To strengthen spatial detail learning and boundary recovery, Sobel operator convolution and dilation operation are performed on the ground truth mask to generate mask pixels that are within distance $d$ from the contours and to use them as the target of auxiliary boundary loss. The procedure is illustrated in Figure \ref{Sobel_Boundary_Formation} and detailed in Algorithm \ref{Algorithm1}. For a segment with length or width less than $2d$, its whole ground truth is displayed in the boundary target mask, such as the small rivers, narrow strips of rangeland, and individual urban building shown in Figure \ref{boundary_targets}. Therefore, all the small objects and segments, which are more likely misjudged, are selected and penalized again by the Sobel Boundary Loss. For any object or segment whose width exceeds $2d$, only its contour of width $d$ is displayed in the boundary target mask, such as the vast agricultural land, large rangeland, and urban residential community shown in Figure \ref{boundary_targets}. By capturing category impurity areas, compared with the ground truth mask, in the boundary target mask, the pixel number ratio of larges segment to small segments drops from quadratic with the segment size ratio to linear with the segment size ratio. In this way, the Sobel Boundary Loss guides the network to learn the features of spatial details.

\begin{algorithm}[H]
\caption{Sobel Boundary Target Generation.}\label{Algorithm1}
\renewcommand{\algorithmicrequire}{\textbf{Input:}}
\renewcommand{\algorithmicensure}{\textbf{Output:}}
\begin{algorithmic}
\REQUIRE Ground Truth $\mathbf{Y}$, Sobel Operator $\mathbf{S_x}$, $\mathbf{S_y}$, Dilation Rate $d$. 
\ENSURE Sobel Boundary Target $\mathbf{\hat{Y}}$. \\
\STATE  
\STATE  $ \mathbf{X_b} \gets (|Conv(\mathbf{Y}, \mathbf{S_x})| + |Conv(\mathbf{Y}, \mathbf{S_y})|) > 0 $ 
\STATE  $ \mathbf{X_d} \gets Dilate(\mathbf{X_b})$ 
\STATE  $\mathbf{\hat{Y}} \gets \mathbf{Y} \otimes \mathbf{X_d}$
\STATE \textbf{return}  $\mathbf{\hat{Y}}$ 
\STATE 
\STATE $\otimes$ means elementwise multiplication.
\end{algorithmic}
\end{algorithm}

$d$ is a hyperparameter that controls the extent of contour pixels participating in the Sobel Boundary Loss calculation. It is not advisable to set $d$ too small for four reasons. Firstly, different from generous images in which objects have clear contours, the land boundaries in land-cover satellite images are comparatively vague. Secondly, setting buffer zones of width $d$ benefit the network learning how to discriminate different categories. Thirdly, if $d$ is set too small, the samples participating in boundary loss calculation are too scarce. Last, the margin of human labeling error should be considered. It is suggested to set $d$ as the value equal to the half of the smaller dimension of most small segments, so the whole bodies of most small segments would remain on the boundary target mask. Without loss of generality, $d$ was set as 50 pixels in the illustration figures and experiments.

\subsection{Total Loss}

\noindent The total loss $L_t$ is the weighted sum of the main semantic segmentation loss $L_m$, auxiliary semantic segmentation losses $L_a$ and boundary losses $L_b$:

\begin{equation}\label{eq:loss}
L_t = w_1 \times L_m + w_2 \times L_a + w_3 \times L_b
\end{equation}

The values of the weights are adjusted according to the values of the loss functions and practical results. If the interiors of large segments are predicted fairly well but the boundaries or small segments are not predicted well, it is advisable to increase $w_3$. To evaluate whether the existence of the auxiliary losses would boost accuracy, without loss of generality, all the weights were set as 1 in the following experiments to avoid any hyperparameter tuning trick.

\section{Experimental Results}
\noindent Since the MKA module specializes in multi-scale feature extraction, to evaluate its effect, we built an image classification network based on the encoder of MKANet and conducted experiments on a scene classification dataset of multi-scale remote sensing images. Then we measured inference speeds of MKANet at various image sizes to validate whether its architecture design can boost inference speed and support large image size. At last, we conducted experiments on two land-cover classification datasets to examine the accuracy of MKANet and the effectiveness of the Sobel Boundary Loss.

\subsection{Image Classification}
\begin{figure}[!t]
\centering
\includegraphics[width=1in]{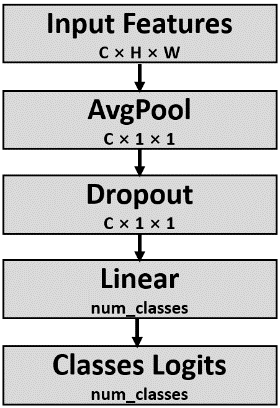}
\caption{Structure of image classification head.}
\label{classification_head}
\end{figure}

\begin{figure}[!t]
\centering
\includegraphics[width=3.5in]{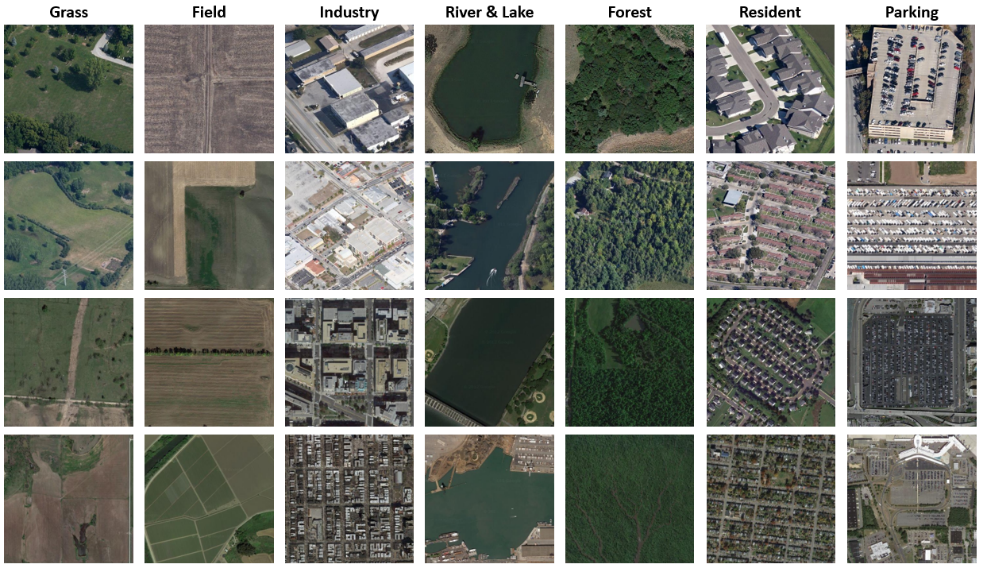}
\caption{Images and categories of the RSSCN7 scene classification dataset.}
\label{RSSCN7}
\end{figure}

\noindent We add an image classification head as illustrated in Figure \ref{classification_head} on top of MKANet encoder for the task of image classification, and name it MKANet-Class. We compared MKANet-Class with other state-of-the-art lightweight classification networks on the RSSCN7 \cite{zou2015deep} scene classification dataset of remote sensing images (Figure \ref{RSSCN7}). RSSCN7 contains seven typical scene categories of image size 400×400 pixels. For each category, 400 images are sampled on four different scales with 100 images per scale. A total of 2800 images were resized to 384×384 pixels and split into a training set, validation set, and test set at a ratio of 2:1:1.

Training details: Cross-entropy was selected as the loss function, and AdamW \cite{loshchilov2017decoupled} was selected as the optimizer with a batch size of 32. The base learning rate was 0.001 with cosine decay. The number of epochs was 150 with a warmup strategy in the first 10 epochs. For a fair comparison, all the networks were trained from scratch without pretraining on other datasets.

Data augmentation: Random flipping, random rotation, and color jittering operations were employed on the input images in the training process.

\begin{table}
\begin{center}
\caption{Comparison of MKANet with other state-of-the-art, efficient classification networks on the RSSCN7 dataset.}
\label{tab_RSSCN7}
\begin{tabular}{ l  c  c }
\hline
\bf{Method} & \bf{Accuracy} & \bf{FPS} \\
\hline
MobileNetV2 \cite{sandler2018mobilenetv2} & 90.43\% & 130.2\\
MobileNetV3 \cite{howard2019searching} & 90.43\% & 107.9 \\
EfficientNet-B0  \cite{tan2019efficientnet} & 91.28\% & 80.1 \\
ShuffleNetV2 x1.0  \cite{ma2018shufflenet} & 91.85\% & 123.5 \\
ResNet18 \cite{he2016deep} & 91.71\% & 178.8 \\
ResNet34 \cite{he2016deep} & 92.14\% & 97.7 \\
STDC1 \cite{fan2021rethinking} & 92.43\% & 143.8 \\
\hline 
MKANet-Class Small & {\bf{92.85\%}} & {\bf{314.3}} \\
\hline 
\end{tabular}
\end{center}
Inference speed frames per second (FPS) were measured on a computer with CPU AMD 4800HS, GPU NVIDIA RTX 2060 Max-Q 6G, and a Pytorch environment.
\end{table}

As shown in Table \ref{tab_RSSCN7}, MKANet-Class Small outperforms other state-of-the-art lightweight classification networks with better accuracy and significantly faster inference speed, which verifies the effectiveness of the MKA module and justifies the efficiency of parallel branch design.

\subsection{Semantic Segmentation Inference Speed} \label{Inference_Speed} 
\noindent For the semantic segmentation task, the inference speeds (FPS) of various networks were measured at 4 image sizes. As shown in Table \ref{tab_FPS}, all the lightweight networks have an obvious advantage over large networks in large image size support and inference speed. On a computer with GPU NVIDIA RTX 3060 12G, none of the large networks can process images with a resolution of 4096×4096 pixels, but all the lightweight networks can. MKANet Small is even capable of processing images up to 7200×7200 pixels and is approximately 2X faster than other lightweight networks and more than 13X faster than the large networks. The large size and fast acquisition speed of satellite images highlight the value of MKANet in accelerating the cognition speed of remote sensing images.

\begin{table*}
\begin{center}
\caption{Inference speed at 4 image sizes; the number of classes is 10.}
\label{tab_FPS}
\begin{tabular}{ l | c  c  c  c  c }
\hline
 & \multicolumn{5}{ c }{\bf{Inference Speed (FPS)}} \\
{\bf{Method}} & \bf{1024×1024} & \bf{2048×2048} & \bf{4096×4096} & \bf{7200×7200} \\ 
\hline
{\bf{Large Networks:}} &  &  &  &  & \\
UNet (VGG16) \cite{ronneberger2015u} & 4.27 (0.6X) & * & * & * \\
PSPNet (ResNet50) \cite{zhao2017pyramid}  & 5.28 (0.7X) & 1.51 (0.7X) & * & * \\
DeepLabV3+ (Xception) \cite{chen2018encoder} & 7.41 (1.0X) & 2.03 (1.0X) & *  & * \\
\hline
{\bf{Lightweight Networks:}} &  &  &  &  \\
ENet \cite{paszke2016enet} & 28.36 (3.8X) & 9.31 (4.6X) & 2.38 (1.0X)  & 0.71 (1.0X) \\
ABCNet \cite{li2021abcnet} & 35.24 (4.8X)  & 10.05 (5.0X) & 2.70 (1.1X)  &  *  \\
BiSeNetV1 (Resnet18) \cite{yu2018bisenet} & 42.94 (5.8X) & 11.85 (5.8X) & 3.23 (1.4X)  &  *  \\
BiSeNetV2 \cite{yu2021bisenet} & 43.22 (5.8X) & 11.74 (5.8X) & 2.99 (1.3X)  &  *  \\
STDC1 \cite{fan2021rethinking} & 52.05 (7.0X) & 15.13 (7.5X) & 4.18 (1.8X)  &  *  \\
STDC2 \cite{fan2021rethinking} & 38.89 (5.2X) & 11.59 (5.7X) & 3.11 (1.3X)  &  *  \\
DDRNet23 \cite{hong2021deep}  & 36.82 (5.0X) & 10.04 (4.9X) & 2.77 (1.2X)  &  0.97 (1.4X) \\
DABNet \cite{li2019dabnet} & 43.89 (5.9X) & 11.54 (5.7X) & 3.05 (1.3X)  &  *   \\
\hline 
MKANet Small  & \bf{98.24 (13.3X)} & \bf{26.98 (13.3X)} & \bf{7.47 (3.1X)}  & \bf{2.41 (3.4X)} \\
MKANet Base  & 59.38 (8.0X) & 16.51 (8.1X) & 4.61 (1.9X) &  *   \\
MKANet Large  & 43.77 (5.9X) & 12.31 (6.1X) & 3.44 (1.5X)  &  *  \\
\hline 
\end{tabular}
\end{center}
* means not executable due to GPU memory overflow.\\
Inference speed frames per second (FPS) were measured on a computer with CPU INTEL i5-3470, GPU NVIDIA RTX 3060 12G, and a Pytorch environment.
\end{table*}


\subsection{Land-cover classification}
\noindent To assess the semantic segmentation performance of MKANet, experiments were conducted on two land-cover classification datasets of satellite images: DeepGlobe Land Cover \cite{demir2018deepglobe} and GID Fine Land Cover Classification \cite{tong2020land}.

The DeepGlobe Land Cover dataset consists of RGB satellite images of size 2448×2448 pixels, with a pixel resolution of 50 cm. The total area size of the dataset is $1716.9km^2$. There are 6 rural land-cover categories. Only the labels of the original training set of the competition have been released; thus, the original training set, which contains 803 images, was split into a training set, validation set, and test set at a ratio of 2:1:1, as described in a previous study \cite{wei2021deanet}.

The GID Fine Land Cover Classification dataset consists of 10 submeter RGB satellite tiles of size 6800×7200 pixels. There are 15 land-cover categories. For the limitation of GPU memory capacity, the 10 tiles were cropped into 90 subimages of size 2400×2400 pixels, and then the 90 subimages were split into a training set, validation set, and test set at a ratio of 3:1:1, similar to a previous study \cite{ding2021looking}.

Training details: For all the lightweight networks, cross-entropy was selected as the loss function, and AdamW \cite{loshchilov2017decoupled} was selected as the optimizer with a batch size of 6, and the base learning rate was 0.001 with cosine decay. The networks were trained for 300 epochs with the DeepGlobe Land Cover dataset and for 500 epochs with the GID dataset, using a warmup strategy in the first 10 epochs. For a fair comparison, all the networks were trained from scratch without pretraining on other datasets.

Data augmentation: Random flipping, random rotation, random scaling of rates (0.7, 0.8, 0.9, 1.0, 1.25, 1.5, 1.75), random cropping into size 1600×1600 pixels, and color jittering operations were employed on the input images during the training process. In the test process, no data augmentation operations were implemented.

Evaluation metrics: The performance of the networks was evaluated by the mean intersection over union (MIoU), and the mean F1 score that are defined as:

\begin{equation}
\label{eq_MIoU}
 \text{MIoU} = \frac {1}{N} \sum_{c=1}^{N} \frac {TP_c}{TP_c + FP_c + FN_c}.
\end{equation}

\begin{equation}
\label{eq_MIoU}
 \text{MF1} = \frac {1}{N} \sum_{c=1}^{N} \frac {2 \times \frac {TP_c}{TP_c + FP_c} \times \frac {TP_c}{TP_c + FN_c}}{\frac {TP_c}{TP_c + FP_c} + \frac {TP_c}{TP_c + FN_c}}.
\end{equation}

where $N$ represents the number of categories and $TP_c$, $FP_c$ and $FN_c$ denote the number of true positive pixels, false positive pixels, and false negative pixels, respectively, in Category $c$.

\subsubsection{DeepGlobe Land Cover Dataset Experimental Results}
\noindent The DeepGlobe Land Cover Classification dataset provides high-resolution sub-meter satellite imagery focusing on rural areas. Due to the variety of land cover types, it is more challenging than the ISPRS Vaihingen and Potsdam datasets \cite{web:2D_Semantic_Labeling} and the Zeebruges dataset \cite{campos2016processing}. The image size of this dataset is so high that only lightweight networks are able to be trained and predict on the original images. We selected this large dataset to evaluate the performance of MKANet on high spatial resolution remote sensing imagery with image size beyond 2K pixels.

As shown in Table \ref{tab_DeepGlobe_efficient} and Table \ref{tab_DeepGlobe_large}, MKANets lead other competitive lightweight networks by at least 3\% and even surpass the large networks with pretrained backbones. In a previous study \cite{wei2021deanet}, to fit the large networks into GPU memory, the authors compressed the images to half size and then divided them into subimages with a resolution of 512×512 pixels. For the large networks, spatial detail loss due to compression offsets their stronger feature extraction ability enabled by a larger backbone, while long-range context information loss due to subdivision weakens their better modeling ability benefitted by a more complex structure. Hence, for large-sized remote sensing patches, lightweight networks have their advantage and can have comparable and even better accuracy than large networks.

\begin{table*}
\begin{center}
\caption{Comparison of MKANets with other state-of-the-art efficient networks on the DeepGlobe Land Cover dataset.}
\label{tab_DeepGlobe_efficient}
\begin{tabular}{ l | c  c  c  c  c  c | c  c }
\hline
 & \multicolumn{6}{ c | }{\bf{Per-category IoU (\%)}} & \bf{MIoU} & \bf{Speed} \\ 
{\bf{Method}} & \bf{Urban} & \bf{Agricult} & \bf{Range} & \bf{Forest} & \bf{Water}  & \bf{Barren} & \bf{(\%)} &  \bf{(FPS)}\\ 
\hline
{\bf{Proportion (\%)}} & 9.35 & 56.76 & 10.21 & 13.75 & 3.74 & 6.14 &\\
\hline 
ENet \cite{paszke2016enet} & 71.14 & 80.86 & 0.42 & 75.44 & 73.10 & 49.41 & 58.39 & 6.96\\
ABCNet \cite{li2021abcnet} & 73.52 & 82.03 & 29.03 & 75.95 & 72.50 & 55.94 & 64.83 & 7.15\\
BiSeNetV1 (Resnet18) \cite{yu2018bisenet} & 72.68 & 83.91 & 28.09 & 78.97 & 74.71 & 56.71 & 65.85 & 8.41\\
BiSeNetV2 \cite{yu2021bisenet} & 72.46 & 82.23 & 23.52 & 77.86 & 69.37 & 51.03 & 62.75 & 8.42\\
STDC1 \cite{fan2021rethinking} & 74.68 & 85.05 & 31.75 & 76.82 & 75.34 & 59.11 & 67.12 & 11.81\\
STDC2 \cite{fan2021rethinking} & 73.48 & 83.79 & 30.57 & 76.27 & 73.48 & 58.11 & 65.95 & 8.76\\
DDRNet23 \cite{hong2021deep} & 75.01 & 84.53 & 32.62 & 77.81 & 77.29 & 58.50 & 67.63 & 7.21\\
DABNet \cite{li2019dabnet} & 74.44 & 84.69 & 33.51 & 79.03 & 75.57 & 58.63 & 67.64 & 8.04\\
\hline 
MKANet Small  & 76.14 & 86.04 & 39.07 & 80.62 & 79.28 & 62.96 & 70.68 & \bf{19.18}\\
MKANet Base  & 76.35 & \bf{87.30} & 39.63 & 81.44 & \bf{81.16} & 64.77 & 71.78
& 11.51\\
MKANet Large  & \bf{76.53} & \bf{87.30} &  \bf{40.36} & \bf{81.66} & 81.04 & \bf{65.08} & \bf{72.00} & 8.59\\
\hline 
\end{tabular}
\end{center}
Inference speed frames per second (FPS) were measured on a computer with CPU INTEL i5-3470, GPU NVIDIA RTX 3060 12G, and a Pytorch environment.
\end{table*}

\begin{table}
\begin{center}
\caption{Comparison of MKANets with large networks on the DeepGlobe Land Cover dataset.}
\label{tab_DeepGlobe_large}
\begin{tabular}{ l  c  c }
\hline
\bf{Method} & \bf{MIoU(\%)} & \bf{MF1(\%)} \\
\hline
UNet (Res2Net50) \cite{ronneberger2015u} & 67.57 & 79.50\\
PSPNet (Res2Net50) \cite{zhao2017pyramid} & 69.45 & 81.07 \\
DeepLabV3 (ResNet50) \cite{chen2017rethinking} & 68.39 & 80.27 \\
DeepLabV3 (ResNet101)  \cite{chen2017rethinking} & 68.94 & 80.55 \\
DeepLabV3+ (Res2Net50)  \cite{chen2018encoder} & 69.12 & 80.85 \\
DeepLabV3+ (Res2Net101)  \cite{chen2018encoder} & 69.39 & 81.06 \\
EncNet (Res2Net50) \cite{zhang2018context} & 68.53 & 80.42 \\
EncNet (Res2Net101) \cite{zhang2018context} & 68.60 & 80.40 \\
PSANet (ResNet50) \cite{zhao2018psanet} & 68.27 & 80.13 \\
GCNet (ResNet50) \cite{cao2019gcnet} & 69.09 & 80.47 \\
DEANet \cite{wei2021deanet} & 71.80 & 82.60 \\
\hline 
MKANet Small & 70.68 & 81.69 \\
MKANet Base & 71.78 & 82.42 \\
MKANet Large & {\bf{72.00}} & {\bf{82.62}} \\
\hline 
\end{tabular}
\end{center}
All the large networks employed the backbones pre-trained on ImageNet \cite{deng2009imagenet}.\\
For all the large networks, the experimental results reported by Wei et al \cite{wei2021deanet} were quoted.
\end{table}

\begin{figure}[!t]
\centering
\includegraphics[width=3.5in]{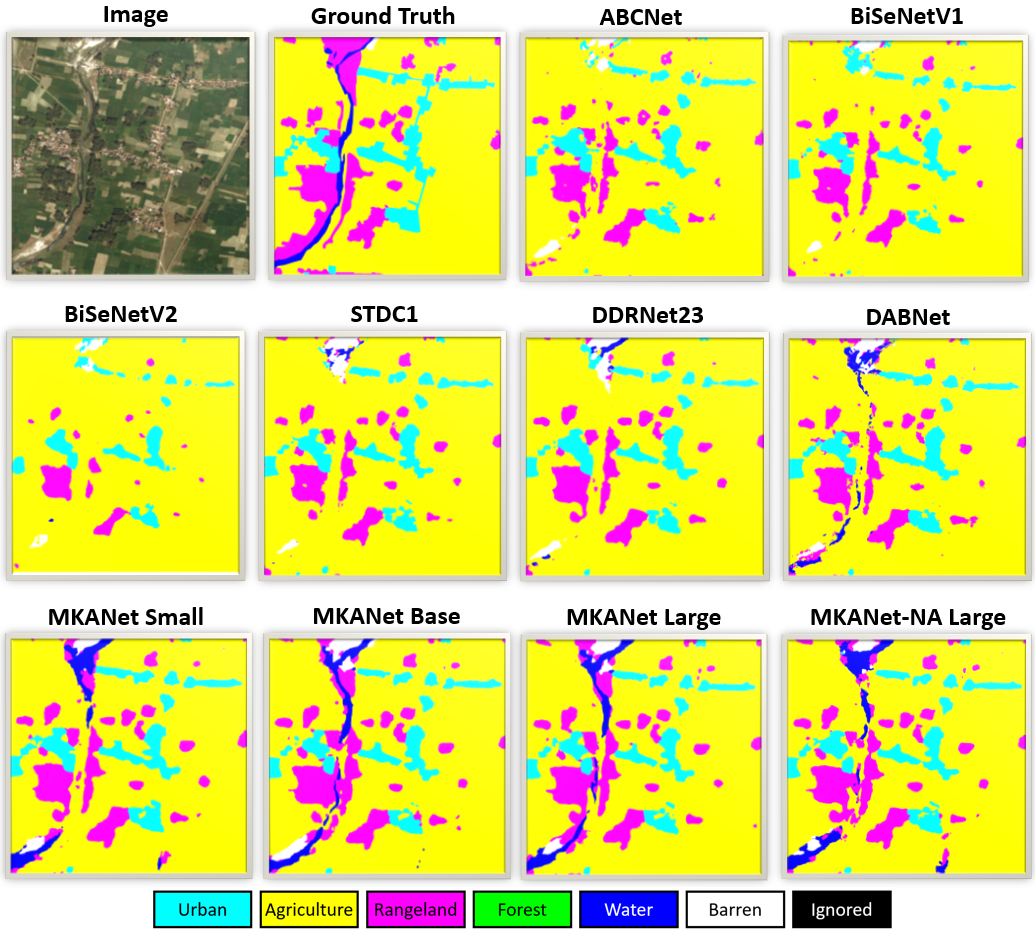}
\caption{Comparison of the masks predicted by different methods on the DeepGlobe land cover dataset.}
\label{predicted_masks}
\end{figure}

As shown in Figure \ref{predicted_masks}, the predicted masks demonstrate that compared with other lightweight networks, MKANets can better identify land cover of small dimensions, for example the river. This superiority is attributed to two factors, one is the MKA module, which has multiscale receptive fields without losing spatial resolution, so spatial details can be preserved. As shown in the lower right subfigure, even without any auxiliary loss, MKANet-NA Large still predicted the river better than other networks. The other factor is the Sobel Boundary Loss, which benefits the network in small segments recognition and boundary recovery. As shown in the bottom of Figure \ref{predicted_masks}, MKANet Base and Large predicted the river more precisely than MKANet-NA Large.

The above observation accords with the per-category IoUs in Table \ref{tab_DeepGlobe_efficient}, MKANets lead other networks by a large margin in the categories of small segments, such as water and range.

\subsubsection{GID Fine Land Cover Classification Dataset Experimental Results}
\noindent The GID Fine Land Cover Classification dataset is very challenging due to its small sample size (only 10 tiles of size 6800×7200 pixels) and highly skewed category distribution, within which the proportions of three categories are as scarce as less than 1\%. We selected this rich category dataset to assess the discrimination ability of MKANet on fine and similar land-cover categories, and also evaluate its robustness with regard to highly unbalanced remote sensing imagery.

As shown in Table \ref{tab_GID_efficient} and Table \ref{tab_GID_large}, MKANets outperform all other lightweight networks and the large networks by a large margin.

\begin{table*}
\begin{center}
\caption{Comparison of MKANets with other state-of-the-art, efficient networks on the GID dataset.}
\label{tab_GID_efficient}
\begin{tabular}{ l |  p{4.5mm}<{\centering}  p{4.5mm}<{\centering}  p{4.5mm}<{\centering}  p{4.5mm}<{\centering}  p{4.5mm}<{\centering}  p{4.5mm}<{\centering}  p{4.5mm}<{\centering}  p{4.5mm}<{\centering}  p{4.5mm}<{\centering}  p{4.5mm}<{\centering}  p{4.5mm}<{\centering}  p{4.5mm}<{\centering}  p{4.5mm}<{\centering}  p{4.5mm}<{\centering}  p{4.5mm}<{\centering} | p{5.5mm}<{\centering} | p{5.0mm}<{\centering}}
\hline
 & \multicolumn{15}{ c | }{\bf{Per-category IoU (\%)}} & \bf{MIoU} & \bf{Speed} \\
{\bf{Method}} & \bf{Indust} & \bf{Urban} & \bf{Rural} & \bf{Traff.} & \bf{Paddy}  & \bf{Irrig.} & \bf{Crop} & \bf{Gard.} & \bf{Arbor} & \bf{Shrub} & \bf{Natur.}  & \bf{Artif.} & \bf{River} & \bf{Lake}  & \bf{Pond} & \bf{(\%)} &  \bf{(FPS)}\\
\hline
{\bf{Proportion (\%)}} & 7.26 & 13.96 & 5.59 & 4.74 & 4.38 & 36.39 & 2.86 & 0.82 & 9.04 & 0.3 & 1.68 & 0.83 & 5.65 & 3.16 & 3.34 &\\
\hline 
ENet \cite{paszke2016enet} & 49.98 & 56.85 & 45.31 & 1.08 & 0.00 & 67.74 & 29.29 & 0.00 & 86.79 & 0.00 & 62.28 & 0.00 & 38.78 & 47.04 & 3.67 & 32.59 & 6.73 \\
ABCNet \cite{li2021abcnet} & 47.09 & 58.88 & 41.21 & 34.01 & 17.47 & 70.32 & 7.29 & 12.05 & 88.96 & 16.48 & 83.43 & 22.50 & 39.74 & 43.10 & 57.85 & 42.69 & 7.20\\
BiSeNetV1 \cite{yu2018bisenet}  & 61.09 & 63.63 & 41.75 & 42.56 & 16.70 & 72.72 & 3.73 & 11.27 & 87.82 & 17.44 & 84.42 & 45.18 & 64.35 & 50.94 & 57.10 & 48.05 & 8.74\\
BiSeNetV2 \cite{yu2021bisenet}  & 59.26 & 65.17 & 49.34 & 43.99 & 56.00 & 79.00 & 3.13 & 9.38 & 93.69 & 9.32 & 86.40 & 55.42 & 39.37 & 44.29 & 50.37 & 49.61 & 8.76\\
STDC1 \cite{fan2021rethinking}  & 62.77 & 64.07 & 46.37 & 52.91 & 13.22 & 73.72 & 4.97 & 9.35 & 69.36 & 3.73 & 86.27 & 56.94 & 86.15 & 75.78 & 59.21 & 50.99 & 12.09\\
STDC2 \cite{fan2021rethinking}  & 57.14 & 60.13 & 44.04 & 45.45 & 29.12 & 72.98 & 13.89 & 11.69 & 87.15 & 6.93 & 87.53 & 44.16 & 62.94 & 54.08 & 54.78 & 48.80 & 9.04\\
DDRNet23 \cite{hong2021deep}  & 63.57 & 65.11 & 50.43 & 59.66 & 53.26 & 80.12 & 11.60 & 9.87 & 84.83 & 0.00 & 87.22 & 63.23 & 86.82 & 75.92 & 53.11 & 56.32 & 7.46\\
DABNet \cite{li2019dabnet}  & 67.52 & 71.08 & 59.64 & 71.10 & 73.86 & 83.79 & 24.44 & 11.72 & 95.48 & 4.15 & \bf{93.27} & 56.99 & 80.38 & 63.41 & 66.73 & 61.57 & 8.37\\
\hline 
MKANet Small   & 67.42 & 69.68 & 55.82 & 70.91 & 70.12 & 82.67 & 37.41 & 16.44 & 93.82 & 10.98 & 92.30 & 61.39 & 86.34 & \bf{78.90} & 69.70 & 64.26 & \bf{19.77}\\
MKANet Base   & \bf{68.82} & \bf{71.39} & 56.83 & 71.92 & \bf{76.18} & 83.43 & 35.57 & 13.40 & 96.00 & 12.49 & 92.22 & 65.08 & 86.70 & 74.46 & 71.93 & 65.09 & 12.01\\
MKANet Large   & 67.90 & 70.35 & \bf{60.00} & \bf{72.88} & 73.31 & \bf{84.50} & \bf{41.20} & \bf{20.76} & \bf{96.75} & \bf{19.66} & 93.22 & \bf{68.77} & \bf{89.30} & 77.37 & \bf{73.49} & \bf{67.30} & 8.86\\
\hline 
\end{tabular}
\end{center}
Inference speed frames per second (FPS) were measured on a computer with CPU INTEL i5-3470, GPU NVIDIA RTX 3060 12G, and a Pytorch environment.
\end{table*}

\begin{table*}
\begin{center}
\caption{Comparison of MKANets with large networks on the GID dataset.}
\label{tab_GID_large}
\begin{tabular}{ l |  p{4.5mm}<{\centering}  p{4.5mm}<{\centering}  p{4.5mm}<{\centering}  p{4.5mm}<{\centering}  p{4.5mm}<{\centering}  p{4.5mm}<{\centering}  p{4.5mm}<{\centering}  p{4.5mm}<{\centering}  p{4.5mm}<{\centering}  p{4.5mm}<{\centering}  p{4.5mm}<{\centering}  p{4.5mm}<{\centering}  p{4.5mm}<{\centering}  p{4.5mm}<{\centering}  p{4.5mm}<{\centering} | p{5.0mm}<{\centering} | p{4.5mm}<{\centering}}
\hline
 & \multicolumn{15}{ c | }{\bf{Per-category F1 Score(\%)}} & \bf{MF1} & \bf{MIoU} \\
{\bf{Method}} & \bf{Indust} & \bf{Urban} & \bf{Rural} & \bf{Traff.} & \bf{Paddy}  & \bf{Irrig.} & \bf{Crop} & \bf{Gard.} & \bf{Arbor} & \bf{Shrub} & \bf{Natur.}  & \bf{Artif.} & \bf{River} & \bf{Lake}  & \bf{Pond} & \bf{(\%)} & \bf{(\%)} \\
\hline
{\bf{Proportion (\%)}} & 7.26 & 13.96 & 5.59 & 4.74 & 4.38 & 36.39 & 2.86 & 0.82 & 9.04 & 0.3 & 1.68 & 0.83 & 5.65 & 3.16 & 3.34 & & \\
\hline 
FCN \cite{shelhamer2016fully} & 68.92 & 73.99 & 64.51 & 68.73 & 74.08 & 84.20 & 68.42 & 24.34 & 87.79 & 4.07 & 53.04 & 25.86 & 83.30 & 66.66 & 77.47 & 62.89 & 49.52 \\
PSPNet \cite{zhao2017pyramid} & 69.18 & 74.41 & 64.87 & 68.09 & 74.53 & 84.69 & 68.23 & 25.26 & 87.84 & 10.36 & 51.87 & 29.07 & 83.15 & 66.71 & 77.40 & 63.57 & 49.98 \\
DeepLabV3+ \cite{chen2018encoder}  & 69.11 & 75.02 & 64.96 & 67.33 & 75.26 & 85.68 & 69.54 & 18.45 & 88.25 & 5.57 & 49.88 & 33.01 & 88.36 & 79.00 & 80.21 & 64.45 & 51.59 \\
DANet \cite{nam2017dual}  & 69.77 & 74.81 & 65.62 & 69.19 & 75.58 & 84.99 & 66.72 & 20.71 & 88.33 & 13.53 & 59.18 & 29.45 & 83.46 & 67.93 & 78.10 & 64.29 & 50.78 \\
SCAttNet \cite{li2020scattnet}  & 68.64 & 73.97 & 64.63 & 64.42 & 71.47 & 85.25 & 70.33 & 22.85 & 87.57 & 3.28 & 56.59 & 24.30 & 86.83 & 73.39 & 77.23 & 63.24 & 50.12 \\
MSCA \cite{zhang2020multi}  & 69.75 & 76.58 & 66.63 & 68.78 & 71.22 & 85.91 & 66.74 & 8.41 & 87.59 & 8.46 & 58.55 & 23.26 & 89.17 & 76.68 & 80.02 & 63.72 & 51.15 \\
LANet \cite{ding2020lanet}  & 69.03 & 75.62 & 65.29 & 68.03 & 72.21 & 85.57 & 67.39 & 7.83 & 88.10 & 10.24 & 54.51 & 30.60 & 87.28 & 74.29 & 78.80 & 63.51 & 50.60 \\
WiCoNet \cite{ding2021looking}  & 69.61 & 75.32 & 65.50 & 67.23 & 73.92 & 86.37 & \bf{72.47} & 31.80 & 88.53 & 13.85 & 47.71 & 42.60 & 87.88 & 76.55 & 81.65 & 65.55 & 52.48 \\
\hline 
MKANet Small   & 80.54 & 82.13 & 71.65 & 82.98 & 82.43 & 90.51 & 54.46 & 28.23 & 96.81 & 19.79 & 96.00 & 76.08 & 92.67 & \bf{88.20} & 82.14 & 74.97 & 64.26 \\
MKANet Base   & \bf{81.53} & \bf{83.30} & 72.48 & 83.67 & \bf{86.48} & 90.96 & 52.47 & 23.63 & 97.96 & 22.21 & 95.95 & 78.84 & 92.88 & 85.36 & 83.68 & 75.43 & 65.09 \\
MKANet Large   & 80.88 & 82.60 & \bf{75.00} & \bf{84.31} & 84.60 & \bf{91.60} & 58.36 & \bf{34.38} & \bf{98.35} & \bf{32.86} & \bf{96.49} & \bf{81.49} & \bf{94.35} & 87.24 & \bf{84.72} & \bf{77.82} & \bf{67.30} \\
\hline 
\end{tabular}
\end{center}
For all the large networks, the experimental results reported by Ding et al. \cite{ding2021looking} were quoted.
\end{table*}

The per-category IoUs in Table \ref{tab_GID_efficient} indicate that the superiority of MKANets is mainly manifested in minor categories, small dimensional categories, and hard-discriminating categories. In these categories, MKANet Small surpasses the average IoU of other lightweight networks by more than 18\%. For example, the large category of farmland (consists of paddy land, irrigated land and dry cropland) is highly skewed in distribution; the samples of irrigated land are approximately 10 times greater than those of paddy land and dry cropland; and MKANet Small exceeds the average IoU of other efficient networks by 37.7\% on paddy land and 25.1\% on dry cropland.

\begin{figure}[!t]
\centering
\includegraphics[width=3.5in]{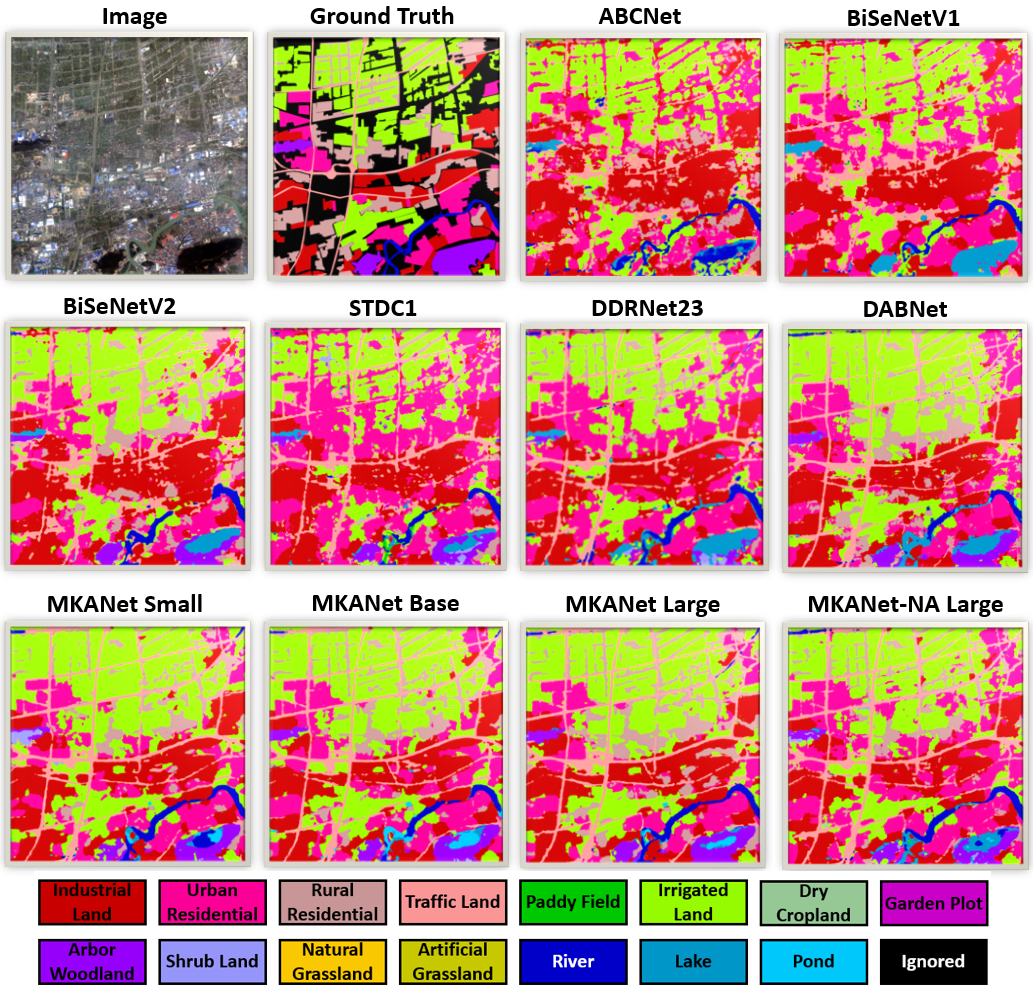}
\caption{Comparison of the masks predicted by different methods on the GID Fine Land Cover Classification dataset.}
\label{GID_predicted_masks}
\end{figure}

As shown in Figure \ref{GID_predicted_masks}, small-dimensional land covers, such as rivers, streets, and rows of rural houses, are better classified by MKANets. MKANet Small exceeds the average IoU of other efficient networks by 27.1\% on traffic land, 18.3\% on artificial grass, 24\% on rivers, 22.1\% on lakes and 19.3\% on ponds.

As shown in the bottom right subfigure, even without any auxiliary loss, MKANet-NA Large still outperformed other lightweight networks in small segments discrimination and spatial detail reconstruction.

\subsection{Ablation analysis}
\noindent To validate the effectiveness of each component, ablation analysis experiments were conducted based on the DeepGlobe Land Cover dataset.

\subsubsection{The influence of input image size on prediction accuracy} \label{ablation various cropping sizes} 
\noindent To investigate the influence of input image size on prediction accuracy, the original DeepGlobe dataset images with a resolution of 2448×2448 pixels were cropped into patches with resolutions of 512×512 pixels, 1024×1024 pixels, and 1600×1600 pixels. The prediction accuracies of these three patch sizes were compared with that of the original image size. As shown in Table \ref{tab_sizes}, the smaller the patch size is, the lower the prediction accuracy is, which demonstrates that the loss of long-range context information caused by cropping images into small patches would incur misjudgement.

\begin{table*}
\begin{center}
\caption{Comparison of various cropping sizes.}
\label{tab_sizes}
\begin{tabular}{ l | c  c  c  c  c  c | c }
\hline
 & \multicolumn{6}{ c | }{\bf{Per-category IoU (\%)}} & \bf{MIoU} \\
{\bf{Method}} & \bf{Urban} & \bf{Agricult} & \bf{Range} & \bf{Forest} & \bf{Water}  & \bf{Barren} & \bf{(\%)} \\
\hline 
MKANet Large,  2448×2448 pixels & 76.53 & 87.30 &  \bf{40.36} & \bf{81.66} & 81.04 & \bf{65.08} & \bf{72.00}\\
MKANet Large,  1600×1600 pixels & \bf{76.78} & \bf{87.38} & 39.84 & 81.36 & 80.92 & 64.50 & 71.80 \\
MKANet Large,  1024×1024 pixels & 76.27 & 86.78 & 38.99 & 80.40 & \bf{81.15} & 64.46 & 71.34 \\
MKANet Large,  512×512 pixels & 76.04 & 85.81 & 36.35 & 79.24 & 79.54 & 59.06 & 69.34 \\
\hline 
\end{tabular}
\end{center}
\end{table*}

\subsubsection{Effectiveness of Kernel-Sharing Atrous Convolution}
\noindent To evaluate the effect of kernel-sharing, atrous convolutions in the MKA module, a variant network with kernel-sharing, atrous convolutions replaced by regular atrous convolutions was built and denoted as MANet. As shown in Table \ref{tab_ablation1}, compared with regular atrous convolutions, kernel-sharing atrous convolutions had stronger feature extraction ability.

\begin{table*}
\begin{center}
\caption{Comparison of networks with kernel-sharing atrous convolutions and regular atrous convolutions.}
\label{tab_ablation1}
\begin{tabular}{ l | c  c  c  c  c  c | c }
\hline
 & \multicolumn{6}{ c | }{\bf{Per-category IoU (\%)}} & \bf{MIoU} \\
{\bf{Method}} & \bf{Urban} & \bf{Agricult} & \bf{Range} & \bf{Forest} & \bf{Water}  & \bf{Barren} & \bf{(\%)} \\
\hline 
MKANet Large  & 76.53 & \bf{87.30} &  \bf{40.36} & \bf{81.66} & 81.04 & \bf{65.08} & \bf{72.00}\\
MANet Large  & \bf{76.87} & 86.88 &  37.63 & 81.09 & \bf{83.42} & 63.68 & 71.59 \\
\hline 
\end{tabular}
\end{center}
\end{table*}

\subsubsection{Effectiveness of Coordinate Attention Module}
\noindent To assess the effect of two coordinate attention modules (CAMs) in the decoder, a variant network with CAMs replaced by simple concatenation operations was built and denoted as MKANet-Concat. As shown in Table \ref{tab_ablation2}, compared with simple concatenation operations, the two CAMs better fused the multiscale features from various stages and augmented the representations of the objects of interest.

\begin{table*}
\begin{center}
\caption{Comparison of MKANets with CAMs and simple concatenation operations.}
\label{tab_ablation2}
\begin{tabular}{ l | c  c  c  c  c  c | c }
\hline
 & \multicolumn{6}{ c | }{\bf{Per-category IoU (\%)}} & \bf{MIoU} \\
{\bf{Method}} & \bf{Urban} & \bf{Agricult} & \bf{Range} & \bf{Forest} & \bf{Water}  & \bf{Barren} & \bf{(\%)} \\
\hline 
MKANet Large  & \bf{76.53} & \bf{87.30} &  \bf{40.36} & \bf{81.66} & 81.04 & \bf{65.08} & \bf{72.00}\\
MKANet-Concat Large  & 76.51 & 86.62 &  37.53 & 80.01 & \bf{81.49} & 61.33 & 70.58 \\
\hline 
\end{tabular}
\end{center}
\end{table*}

\subsubsection{Effectiveness of Auxiliary Losses }
\noindent To estimate the contribution of auxiliary semantic segmentation loss and auxiliary boundary loss, a variant network without any auxiliary loss was built and denoted as MKANet-NA, and another variant network without auxiliary boundary loss was built and denoted as MKANet-NB. As shown in Table \ref{tab_ablation31}, on the DeepGlobe Land Cover dataset, auxiliary semantic segmentation loss improved the MIoU by nearly 1\%, and auxiliary boundary loss further boosted the MIoU by approximately 1.5\%. On the GID Fine Land Cover Classification dataset, both improvements were broadened to 1.9\% (Table \ref{tab_ablation32}). The IoUs of most categories were boosted by the two kinds of auxiliary losses, indicating that these auxiliary heads can promote spatial detail learning at lower levels and semantic context learning at higher levels.

\begin{table*}
\begin{center}
\caption{Comparison of MKANets with and without auxiliary losses on the DeepGlobe Land Cover dataset.}
\label{tab_ablation31}
\begin{tabular}{ l | c  c  c  c  c  c | c }
\hline
 & \multicolumn{6}{ c | }{\bf{Per-category IoU (\%)}} & \bf{MIoU} \\
{\bf{Method}} & \bf{Urban} & \bf{Agricult} & \bf{Range} & \bf{Forest} & \bf{Water}  & \bf{Barren} & \bf{(\%)} \\
\hline 
MKANet Large  & \bf{76.53} & \bf{87.30} &  \bf{40.36} & 81.66 & \bf{81.04} & \bf{65.08} & \bf{72.00}\\
MKANet-NB Large  & 75.71 & 87.25 &  35.51 & \bf{81.83} & 80.51 & 62.19 & 70.50 \\
MKANet-NA Large  & 75.97 & 86.48 &  34.22 & 80.14 & 79.50 & 60.99 & 69.55 \\
\hline 
\end{tabular}
\end{center}
\end{table*}

\begin{table*}
\begin{center}
\caption{Comparison of MKANets with and without auxiliary losses on the GID Fine Land Cover Classification dataset.}
\label{tab_ablation32}
\begin{tabular}{ l |  p{4.5mm}<{\centering}  p{4.5mm}<{\centering}  p{4.5mm}<{\centering}  p{4.5mm}<{\centering}  p{4.5mm}<{\centering}  p{4.5mm}<{\centering}  p{4.5mm}<{\centering}  p{4.5mm}<{\centering}  p{4.5mm}<{\centering}  p{4.5mm}<{\centering}  p{4.5mm}<{\centering}  p{4.5mm}<{\centering}  p{4.5mm}<{\centering}  p{4.5mm}<{\centering}  p{4.5mm}<{\centering} | p{4.5mm}<{\centering} }
\hline
 & \multicolumn{15}{ c | }{\bf{Per-category IoU (\%)}} & \bf{MIoU} \\
{\bf{Method}} & \bf{Indust} & \bf{Urban} & \bf{Rural} & \bf{Traff.} & \bf{Paddy}  & \bf{Irrig.} & \bf{Crop} & \bf{Gard.} & \bf{Arbor} & \bf{Shrub} & \bf{Natur.}  & \bf{Artif.} & \bf{River} & \bf{Lake}  & \bf{Pond} & \bf{(\%)} \\
\hline 
MKANet Large & 67.90 & 70.35 & \bf{60.00} & \bf{72.88} & 73.31 & \bf{84.50} & \bf{41.20} & \bf{20.76} & \bf{96.75} & \bf{19.66} & \bf{93.22} & 68.77 & \bf{89.30} & 77.37 & \bf{73.49} & \bf{67.30} \\
MKANet-NB Large & \bf{69.78} & \bf{71.11} & 57.96 & 70.83 & \bf{73.90} & 82.78 & 39.63 & 13.49 & 96.70 & 15.60 & 91.38 & \bf{69.30} & 84.27 & 72.38 & 71.63 & 65.38 \\
MKANet-NA Large & 69.20 & 69.31 & 53.34 & 69.70 & 70.35 & 82.32 & 39.75 & 10.99 & 92.09 & 2.98 & 90.48 & 64.59 & 88.96 & \bf{82.63} & 65.88 & 63.51 \\
\hline 
\end{tabular}
\end{center}
\end{table*}

\begin{figure}[!t]
\centering
\includegraphics[width=3.5in]{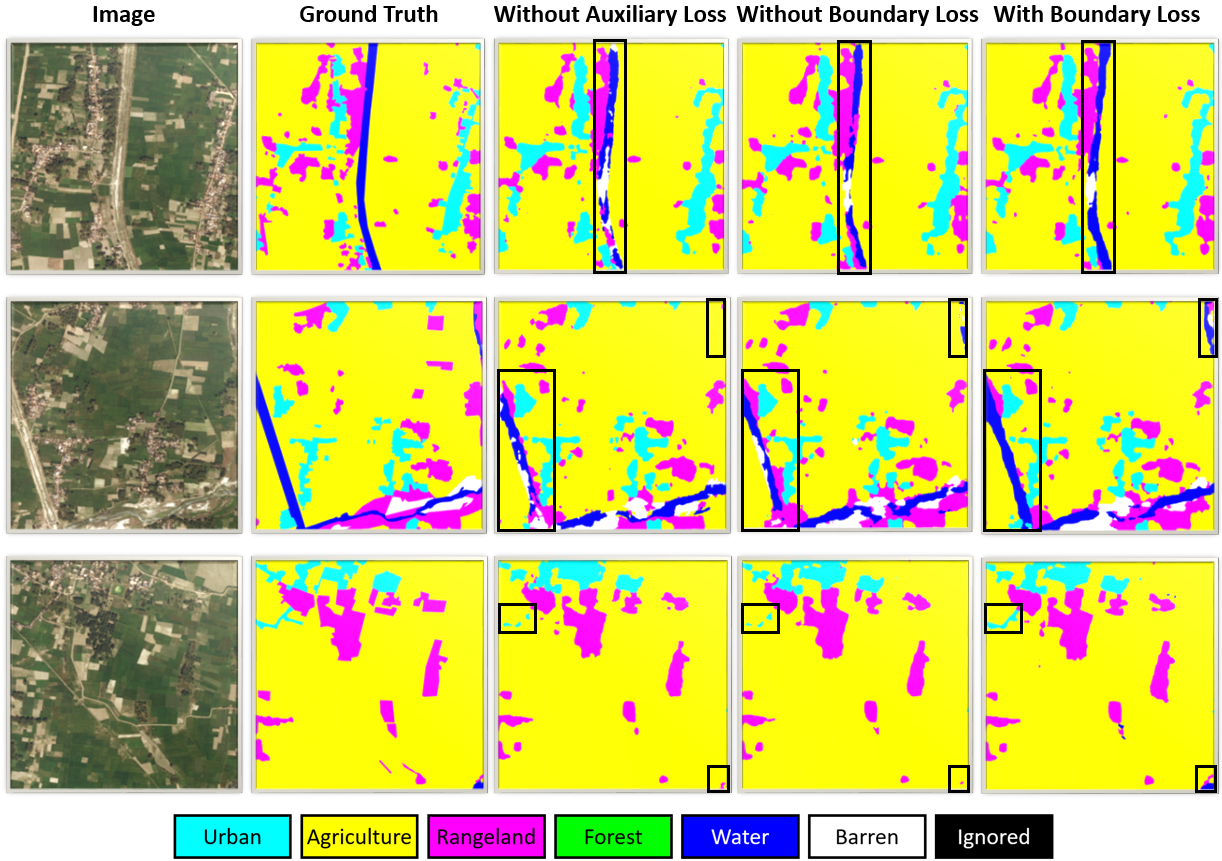}
\caption{Comparison of the masks predicted by MKANets without any auxiliary loss, without auxiliary boundary loss, and with both types of auxiliary losses.}
\label{ablation_boundary}
\end{figure}

To visualize the effect of auxiliary semantic segmentation loss and auxiliary boundary loss, the predicted labels of the above three networks are displayed in Figure \ref{ablation_boundary}. The results showed that the auxiliary losses, especially the boundary loss, can guide the networks to better recognize small segments and restore boundaries, which is in accordance with our design.

\subsubsection{Stacking More MKA Modules per Stage}
\noindent MKANet can be expanded not only in width by increasing the number of base channels $c$ but also in depth by increasing the repeating times $r$ of MKA modules in each stage. As shown in Table \ref{ablation_boundary4}, the performance of MKANet increased by stacking additional MKA modules in each stage.

\begin{table*}
\begin{center}
\caption{Comparison of MKANets with different widths and depths.}
\label{ablation_boundary4}
\begin{tabular}{ l | c  | c }
\hline
 &   \bf{MIoU}  & {\bf{Speed}} \\
{\bf{Method}} &  \bf{(\%)} & \bf{FPS} \\
\hline
MKANet $c = 64, r = 1$ &  70.68  & \bf{19.18}\\
MKANet $c = 64, r = 2$ & 71.70 & 14.22\\
MKANet $c = 96, r = 1$ & 71.78 & 11.51\\
MKANet $c = 96, r = 2$ & 71.84 & 8.47\\
MKANet $c = 128, r = 1$ & \bf{72.00} & 8.59\\
\hline 
\end{tabular}
\end{center}
Inference speed frames per second (FPS) were measured on a computer with CPU INTEL i5-3470, GPU NVIDIA RTX 3060 12G, and a Pytorch environment.
\end{table*}

\subsubsection{The optimal value for the number of branches}

\begin{table*}
\begin{center}
\caption{Comparison of MKANets with a variable number of parallel branches on the DeepGlobe Land Cover dataset.}
\label{tab_ablation5}
\begin{tabular}{ l | c  c  c  c  c  c | c }
\hline
 & \multicolumn{6}{ c | }{\bf{Per-category IoU (\%)}} & \bf{MIoU} \\
{\bf{Method}} & \bf{Urban} & \bf{Agricult} & \bf{Range} & \bf{Forest} & \bf{Water}  & \bf{Barren} & \bf{(\%)} \\
\hline 
MKANet $b = 2, c = 96, r = 1$  & 76.65 & 87.22 &  36.79 & 80.86 & \bf{81.95} & 63.99 & 71.24\\
MKANet $b = 3, c = 96, r = 1$  & 76.35 & \bf{87.30} &  \bf{39.63} & \bf{81.44} & 81.16 & 64.77 & \bf{71.78} \\
MKANet $b = 4, c = 96, r = 1$  & \bf{76.71} & 87.20 &  38.51 & 81.31 & 81.77 & 63.92 & 71.57 \\
MKANet $b = 5, c = 96, r = 1$  & 76.39 & 87.17 &  37.10 & 81.01 & 81.92 & \bf{65.15} & 71.46 \\
\hline 
\end{tabular}
\end{center}
\end{table*}

\noindent In the MKA module, the number of parallel branches $b$ is a hyperparameter, which is proportional to the receptive field and computational cost. To determine the optimal value for $b$, different values were tested. As shown in Table \ref{tab_ablation5}, with one more branch of a larger dilation rate atrous convolution to sense larger-scale features, MKANet $b = 3$ performed better than MKANet $b = 2$ by 0.54\% on the MIoU metric. However, as more branches were added, the MIoU dropped. This negative effect was mainly attributed to Part 3 of the MKA module, where the output features of each branch were concatenated and a pointwise convolution was then applied to them to compress the channels to $1/ b$. The larger $b$ is, the more information losses there are in the channel compression process. $b = 3$ strikes a good balance among the multiscale receptive field, information loss, and computation efficiency; thus, $b$ defaults to the optimal value 3 in the MKA module.

\section{Conclusion}
\noindent Aim at the characteristics of multi-scales and large image size of top view remote sensing imagery, we merged the specialized multi-branch module and the shallow architecture design into MKANet. Through extensive experiments and ablation analysis, MKANet has reached our initial expectation and alleviated the three problems (slow inference speed, incapability of pro-
cessing large size image patches, and easy misjudgement on boundaries and small segments) that restrict the practical applications of semantic segmentation networks in remote sensing imagery.

In the land-cover classification experiments, the original images of 2448×2448 pixels and large patches of 2400×2400 pixels were employed as the input of MKANet. Compared with a cropping size of 512×512 pixels or even smaller in most existing studies, the number of subimages was reduced to 1/25. In the inference speed experiments, MKANet Small can support even larger image size of 7200×7200 pixels on RTX 3060 and image size up to 10k×10k pixels on RTX3090. Its friendly support of large subimages greatly alleviates spatial detail loss due to down sampling or long-range context information loss due to cropping. Meanwhile, MKANet Small is approximately 2X faster than other lightweight networks and 13X faster than the large networks. Both merits highlight the value of MKANet in accelerating the perception and cognition speed of remote sensing imagery.

In response to the problem that prediction errors are more likely to occur on boundaries and small segments, the Sobel operators convolution and dilation operation were innovatively utilized to capture category impurity areas, exploit boundary information and exert an extra penalty on boundaries and small segment misjudgment. Both quantitative metrics and visual interpretations have verified that the Sobel Boundary Loss can promote spatial detail learning and boundary reconstruction.

For the task of land-cover classification, MKANet has successfully raised the benchmark on accuracy and demonstrated that if lightweight efficient networks are properly designed, they can have comparable accuracy with that of large networks. In addition, due to the merits of fast inference speed and a low requirement on hardware, lightweight networks have immense potential in practical applications and are equally important with large networks. Notably, MKANet outperformed other state-of-the-art lightweight networks with significantly better accuracy. 

In future research, we will extend the application of the MKA module as an effective and swift feature extractor in instance segmentation, change detection, and object detection tasks.

\section*{Acknowledgments}
This work was supported by the National Natural Science Foundation of China (No. 61901307), Open Research Fund of State Key Laboratory of Information Engineering in Surveying, Mapping and Remote Sensing, Wuhan University (No. 20E01), Scientific Research Foundation for Doctoral Program of Hubei University of Technology (No. BSQD2020054, No. BSQD2020055).

\begin{IEEEbiography}[{\includegraphics[width=1in,height=1.25in,clip,keepaspectratio]{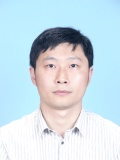}}]{Zhiqi Zhang}
received a B.Sc. degree in Geographic Information Systems from Huazhong Agricultural University, a B.Eng. degree in Computer Science and Technology from Huazhong University of Science and Technology, an M.Eng. degree in Computer Technology from Wuhan University, and a Ph.D. degree in photogrammetry and remote sensing from Wuhan University in 2006, 2006, 2015, and 2018. He is currently an Associate Professor with the School of Computer Science, Hubei University of Technology. His research interests include system architecture, algorithm optimization, AI, and high-performance processing of remote sensing.
\end{IEEEbiography}

\begin{IEEEbiography}[{\includegraphics[width=1in,height=1.25in,clip,keepaspectratio]{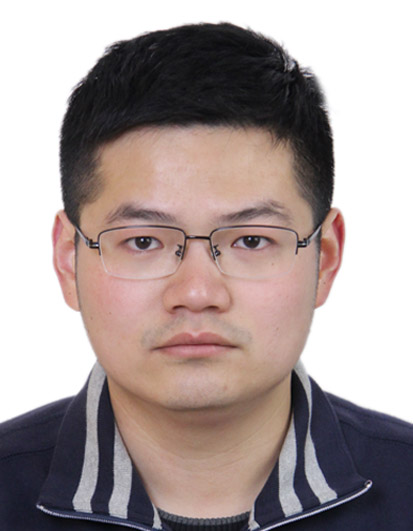}}]{Wen Lu}
is a postgraduate student majoring in Computer Science and Technology in the School of Computer Science, Hubei University of Technology. 
His research interests include computer vision, remote sensing, machine learning, and deep learning.
\end{IEEEbiography}

\begin{IEEEbiography}[{\includegraphics[width=1in,height=1.25in,clip,keepaspectratio]{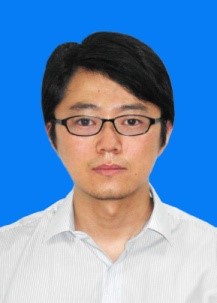}}]{Jinshan Cao}
received a Ph.D. degree in photogrammetry and remote sensing from the School of Remote Sensing and Information Engineering, Wuhan University, Wuhan, China, in 2012.
He is currently an Associate Professor with the School of Computer Science, Hubei University of Technology, Wuhan, China. His research interests include geometric calibration, sensor orientation, and image registration of high-resolution satellite imagery.
\end{IEEEbiography}

\begin{IEEEbiography}[{\includegraphics[width=1in,height=1.25in,clip,keepaspectratio]{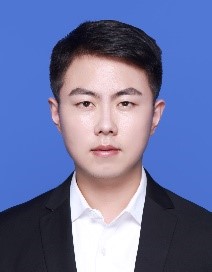}}]{Guangqi Xie}
received an M.Sc. degree in Remote Sensing of Resources and Environment from China University of Geosciences, Wuhan, China, in 2018. He is currently working toward a Ph.D. degree in the State Key Laboratory of Information Engineering in Surveying, Mapping, and Remote Sensing, Wuhan University, Wuhan, China.
His research interests include image matching and registration, pansharpening, and image superresolution.
\end{IEEEbiography}

\bibliographystyle{IEEEtran}
\bibliography{IEEEabrv,mylib}

\end{document}